\documentclass[runningheads]{llncs}

\usepackage[T1]{fontenc}
\usepackage{graphicx}
\usepackage{todonotes}
\usepackage{booktabs}
\usepackage{supertabular}
\usepackage{adjustbox}
\usepackage{float}
\usepackage{tabularx}
\usepackage{tablefootnote}
\usepackage{verbatim}
\usepackage{listings}
\usepackage{hyperref}
\usepackage{lscape}
 \usepackage{amsmath}
 \usepackage{tabularx}
\usepackage{amssymb}
\usepackage{paralist}

\usepackage[inline]{enumitem}

\newcommand{\csets}[2]{\texttt{#1}$\leftrightarrow$\texttt{#2}}

\newcommand{\revont}{\texttt{RevOnt}}
\newcommand{\claro}{\texttt{CLaRO}}
\newcommand{\ontochat}{\texttt{OntoChat}}
\newcommand{\gpt}{\texttt{GPT 4.1}}
\newcommand{\gemini}{\texttt{Gemini 2.5 Pro}}

\renewcommand{\paragraph}[1]{\textbf{#1}\ }
\begin{document}
%
\title{A Comparative Study of Competency Question Elicitation Methods from Ontology Requirements}

\titlerunning{A comparative analysis of CQ elicitation methods}
%
\author{
Reham Alharbi\inst{1}\orcidID{0000-0002-8332-3803} \and
Valentina Tamma\inst{2}\orcidID{0000-0002-1320-610X} \and
Terry R. Payne\inst{2}\orcidID{0000-0002-0106-8731} \and
Jacopo de Berardinis\inst{2}\orcidID{0000-0001-6770-1969}
}

\authorrunning{R. Alharbi et al.}

\institute{
College of Computer Science and Engineering, Taibah University, Saudi Arabia \\
 \email{rfalharbi@taibahu.edu.sa}
\and
School of Computer Science and Informatics, University of Liverpool, UK  \\
\email{\{v.tamma,t.r.payne,jacodb\}@liverpool.ac.uk} 
}
%
%

%
\maketitle              
\begin{abstract}
Competency Questions (CQs) are pivotal in knowledge engineering, guiding the design, validation, and testing of ontologies.
A number of diverse formulation approaches have been proposed in the literature, ranging from completely manual to those that use Large Language Model (LLM).
However, attempts to characterise the outputs of these approaches and their systematic comparison are scarce.
This paper presents an empirical comparative evaluation of three distinct CQ formulation approaches: \emph{manual formulation} by ontology engineers, instantiation of \emph{CQ patterns}, and generation using state of the art \emph{LLMs}.
CQs are generated using each approach given a set of requirements for cultural heritage, and they are assessed empirically across different dimensions: degree of acceptability, ambiguity, relevance, readability and complexity.
Our contribution is twofold: (i) the first multi-annotator dataset of CQs generated from the same source using different methods; and (ii) a systematic comparison of the characteristics of the CQs resulting from each approach.
Our study shows that different CQ generation approaches have different characteristics and that LLMs can be used as a way to initially elicit CQs, however these are sensitive to the model used to generate CQs and they generally require a further refinement step.

\end{abstract}

\section{Introduction}\label{sec:intro}

Competency Questions (CQs)~\cite{gruninger1995methodology} are fundamental in the ontology engineering lifecycle, as they can bridge the gap between domain experts and knowledge engineers. Their primary function is to support the elicitation of the requirements, expressed in the form of questions in natural language, that an ontology should satisfy. CQs scope the knowledge modelled in an ontology and guide \textit{ontology design} activities by suggesting relevant concepts and relations to model~\cite{Presutti2009eXtremeDW,SurezFigueroa2015NEON}. CQs are also instrumental in other phases of the ontology construction process, as they support: ontology \emph{validation and testing} by providing verifiable queries the ontology must answer~\cite{bezerra2017verifying,Keettestdriven2016}; and \emph{ontology reuse} by clarifying the intended scope and capabilities of an existing ontology fragment~\cite{Aziz2023,FernandezLopez2019}.

Despite their importance ~\cite{noy2001ontology,POVEDAVILLALON2022LOT,SurezFigueroa2015NEON}, CQs are often underused in practice as developers find them difficult or too time-consuming to formulate, and lament the lack of standardised tool support~\cite{alharbi2021,monfardini2023}.
A further challenge to their effective adoption  lies in the absence of clear, objective criteria to assess whether the generated CQs appropriately reflect the intended requirements.
As a result, practitioners often struggle to produce questions that are consistently clear, relevant, and suited to the task \cite{keet2024:discerning}. 
To facilitate CQ formulation and uptake, several studies have explored their (semi)-automatic generation, leveraging controlled natural language, e.g., \claro  \cite{CLARO2019}, patterns and archetypes~\cite{bezerra2017verifying,Ren2014,wisniewski2019analysis}, and more recently, Large Language Models (LLMs)~\cite{alharbi2024SAC,AgoCQ_Keet,RevOnt_2024,rebboud2024_ESWC,Bohui2025}.
However, no comprehensive studies exist that systematically assess the efficacy of different generation methods under different conditions, user profiles and stakeholders (i.e. domain experts or ontology engineers) 
within the broader ontology development process.

In this paper, we perform a comparative analysis of three state of the art approaches by systematically analysing the resulting CQs:
(i) \textbf{manual} formulation by experienced ontology engineers; (ii) semi-automated formulation via instantiation of CQ \textbf{patterns} \cite{Ren2014}; (iii) \textbf{automated} formulation using LLMs (\gpt, \gemini).
Each approach is used with a common set of requirements extracted from the same user story within the cultural   heritage domain.
The analysis includes both a qualitative multi-annotator evaluation and a quantitative assessment to characterise the produced CQs in terms of their structural and linguistic features (\textbf{readability} and \textbf{ambiguity}, together with \textbf{complexity} and their \textbf{relevance} to the user story with respect to the underlying requirements).
%
%
%
The paper's contribution is twofold:
\begin{itemize}[itemsep=1pt, topsep=2pt, partopsep=0pt]
   \item A multi-faceted comparative analysis of the CQs comprising:
   \begin{inparaenum}
   \item[(i)] the evaluation of CQ suitability, manually performed by ontology engineers; 
   \item[(ii)] CQ feature extraction and analysis, to characterise the CQs generated by different approaches; and finally, 
   \item[(iii)] the semantic overlap via embeddings to assess to what extent different approaches generate similar CQs.
   \end{inparaenum}
   \item The first multi-annotator dataset of CQs generated from identical source material expressing ontological requirements from the user stories and using distinct elicitation approaches (manual, pattern-based, LLM-based).
\end{itemize}

This is one of the first comparative analysis of manual, pattern-based, and LLM-based approaches from a controlled source (a single user story). We intentionally fixed the domain and source text to run a controlled comparison and isolate the effect of the CQ generation method. Our evaluation includes: (i) manual expert assessment; (ii) ambiguity annotation; and (iii) feature extraction across five datasets. These steps are sensitive to domain-specific terminology and narrative complexity, so we needed a single fixed reference input (the user story). 
Our results show that CQ formulation approaches have distinct characteristics: human formulated CQs generate the highest agreement while producing the most readable and least complex questions, whereas LLM based approaches generate lower agreement, require more knowledge to be understood and are more complex, generally requiring further refinements.

Section~\ref{sec:related} discusses related work, whilst the methodology used for the comparative study (the research questions, experimental design, and evaluation framework) are detailed in Section~\ref{sec:methodology}.
We present the results in Section~\ref{sec:results} and discuss key findings and limitation in Section~\ref{sec:discussion}, before concluding in Section~\ref{sec:conclusions}.
\section{Related work}\label{sec:related}

Competency Questions (CQs)~\cite{uschold1995towards} are a core component of many ontology development methodologies and frameworks~\cite{noy2001ontology,POVEDAVILLALON2022LOT,SurezFigueroa2015NEON},
supporting a range of ontology engineering activities, including scoping the ontology, guiding the reuse of existing ontological resources, informing the naming of classes and relations, and validating the resulting ontological artefacts~\cite{bezerra2017verifying,Dennis2017,Keettestdriven2016,Kim2007}. 
Although they are widely proposed as a means of eliciting domain knowledge and aligning ontological models with stakeholder requirements, their practical application remains inconsistent
\cite{alharbi2021,monfardini2023}, largely due to the lack of systematic methodological support for their formulation.
Existing guidance is often fragmented and fails to clearly distinguish among the various purposes, stages, and quality criteria associated with CQ development.
Despite these limitations, CQs continue to be regarded as a valuable asset for supporting ontology design and validation~\cite{alharbi-et-al2024:ekaw,keet2024:discerning}.

Various approaches have recently emerged to support users in eliciting CQs: 
Ren et.al.  propose a method to define ontology authoring tests through the definition of 
patterns (archetypes) that generate answerable CQs~\cite{Ren2014}.
Wi{\'s}niewski et. al. also proposed instantiating patterns 
with domain knowledge to assist ontology engineers in formulating machine-processable CQs that can be used for ontology testing~\cite{wisniewski2019analysis}.
Bottom-up approaches have  been proposed that use patterns for the generation of Controlled Natural Language (CNL) templates for authoring ontologies \cite{antia-keet-2021-assessing,CLARO2019}.
However, both patterns and templates still require significant effort by ontology developers since 
they need to be contextualised to the different scenarios \cite{antia-keet-2021-assessing}. 
%
%
The use of generative AI has further broadened the landscape of CQ authoring;
recent approaches leverage LLMs and external knowledge sources to support automatic CQ generation, possibly in conversational settings involving domain experts and ontology engineers.
Examples include \ontochat \cite{Bohui2025}, LLM-driven methods for CQ generation, e.g. the work of Rebboud, Y., et al.~\cite{rebboud2024_ESWC}  \revont \cite{RevOnt_2024}, and \texttt{RetrofitCQ} \cite{alharbi2024SAC}.

Despite this progress, a gap remains in the evaluation of (semi-)automatically generated CQs.
Current methods vary significantly in their assessment strategies; some rely on similarity metrics \cite{alharbi2024SAC,rebboud2024_ESWC}, while others incorporate human judgment \cite{AgoCQ_Keet,RevOnt_2024}.
However, the scope of human evaluation is often narrow; for example, \cite{RevOnt_2024} focuses primarily on readability, whereas \cite{AgoCQ_Keet} addresses syntactic correctness, answerability, domain relevance, and coverage.
%
Often these measures have been proposed to assess
question generation approaches, typically complementing others e.g. fluency, clarity, conciseness, and complexity \cite{fu-etal-2024-qgeval,DBLP:journals/pai/MullaG23,ousidhoum-etal-2022-varifocal}.
These evaluation metrics depend on the availability of ground truth, i.e. benchmarks of CQs formulated by ontology engineers for the construction of one or more ontologies~\cite{de2023polifonia,coral}. Yet recent studies~\cite{alharbi2024SAC,keet-kahn2024:ekaw,POTONIEC2020105098}
have shown that even these benchmarks can suffer from poor or ambiguous phrasing, typos, or more fundamental issues such as unanswerable CQs. 

\section{Comparative analysis of CQ elicitation approaches}\label{sec:methodology}
We proposed a framework to compare selected CQ elicitation approaches (manual, pattern-based, and LLM-based).
The CQs in this study are all derived from the same user story-based requirements, therefore eliminating the risk of inconsistencies in the requirements. We investigate the following research questions:
\begin{description}[topsep=2pt]
    \item[\textbf{RQ1:}] To what extent do ontology engineers achieve consensus regarding the suitability of CQs when representing the requirements of a given user story?
    \item[\textbf{RQ2:}] To what extent do CQs formulated by different approaches compare in terms of \textbf{ambiguity}, \textbf{relevance}, \textbf{readability}, and \textbf{complexity}?
    \item[\textbf{RQ3:}] To what extent do CQ sets generated by different approaches compare in terms of diversity and semantic overlap (shared meaning)?
\end{description}


In order to systematically address these research questions, we need to ensure that the CQ elicitation approaches were applied to the same source material. We therefore curated \textbf{AskCQ} -- a novel multi-annotator dataset that comprises 204 CQs generated by distinct CQ formulation approaches applied to the same ontology requirements. The CQs in AskCQ are elicited from a user story developed in a cultural heritage project.
The story involves two personas: a music archivist and a collection curator, and focuses on their interactions, goals, and information needs related to the museum's collection and data management -- from loaning music memorabilia to their metadata, multimedia, and display requirements.
We follow the same user story template from~\cite{de2023polifonia} to include goals, scenarios, example data, in addition to preliminary exploratory questions from the personas.
%
We utilised 3 distinct approaches to elicit CQs from the user story.

\begin{enumerate}[topsep=2pt]
    \item \textbf{Manual.} Two human annotators (\texttt{HA-1}, \texttt{HA-2} with at least 5 years of experience in OE) independently read the user story.
    The annotators were explicitly instructed to formulate CQs that they considered necessary for modelling the \emph{scope} and \emph{functionalities} described in the user story, without any specific constraints on the number, phrasing, or style of the questions.
    \item \textbf{Pattern-based.} This set was generated by another ontology engineer with at least five years of experience in requirements engineering, using a predefined set of CQ templates (archetypes) from \cite{Ren2014}. These templates (e.g., “Which [CE1] [OPE] [CE2]?'', “Find [CE1] with [CE2]'', “Which are [CE]?'') were manually instantiated based on the entities, relations, and information needs identified in the user story.
    We generated CQs from the user story using all patterns proposed by Ren et al.~\cite{Ren2014}. Most patterns could be instantiated, but four main patterns 
    produced no CQs because their structure does not match the entities or relations present in the user story.
    \item \textbf{LLM-based.} These CQs were generated by two state-of-the-art LLMs (\gpt, \gemini) from a markdown version of the user story. Notably, we elected not to provide any definition, number, or desired (and potentially subjective) property of CQs in the prompt in order to avoid biasing in any way the resulting formulation.
\end{enumerate}

The CQs in AskCQ are organised in five distinct sets: HA-1 (44), HA-2 (54), Pattern (38), \gpt (26), and \gemini (42).
All generated CQs are assigned unique identifiers, and anonymised with respect to their generation method to ensure the evaluators were not biased in the subsequent evaluation phases. 
To the best of our knowledge, AskCQ is the first publicly available multi-annotator CQ dataset derived from manual, pattern-based, and LLM-based approaches applied to the same ontological requirements.
The dataset and the user story are both publicly released under a CC-BY licence.


\subsection{CQ-level suitability evaluation and feature extraction}\label{sssec:eval_framework}

The RQs outlined in Section~\ref{sec:methodology} are addressed using a mixed-methods evaluation approach. 
For RQ1, we conducted an expert evaluation with ontology engineers to assess the suitability of each CQ within the user story context.
For RQ2, both manual and NLP-based methods are used to extract and compare relevant CQ-level features.
For RQ3, a semantic analysis based on sentence embeddings was employed to investigate the semantic overlap among the CQ sets.

Each anonymised CQ was independently assessed by three annotators (N=3) with expertise in OE.
Annotators were first required to review the user story and persona descriptions before being assigned to an evaluation pool. 
To mitigate potential bias, the evaluators were distinct from the individuals who manually generated the CQs for the same set.
They were instructed to either accept or reject each CQ
based on its suitability for guiding OE tasks for the given user story.
No specific quality criteria were provided, thereby safeguarding the annotators' independent judgment and minimising evaluative bias, analogously to the way the elicitation process was conducted.
Annotators were also given the opportunity to provide brief comments, particularly to justify rejections or to highlight specific concerns regarding a CQ.
Comments that indicated ambiguity were subsequently identified through manual analysis and used to trigger further discussions and inform the corresponding evaluation metric.
Ultimately, we employed this manual evaluation of \textit{suitability} to capture the expert judgment of ontology engineers regarding domain correctness, ensuring the CQs were grounded in practical modelling requirements.

As there is no consensus on a single definition of CQ quality, we selected a multi-faceted set of metrics -- \emph{ambiguity}, \emph{relevance} to the user story, \emph{readability}, and \emph{complexity} -- to characterise the CQs from complementary perspectives and enable their comparison.
Each metric is described below:


\noindent \paragraph{Ambiguity:} measured as a binary label attributed to a CQ based on the comments contributed in the expert evaluation, as CQs that were marked as ambiguous can still be considered suitable for the OE task.
The same annotators who decided on the suitability reviewed these comments.
Those CQs that received at least one comment triggered a discussion amongst the annotators, since the presence of the comment was interpreted as being indicative of an issue.
After the discussion, the annotators finalised their suitability score of the CQ.
Therefore, we operationalised ``Ambiguity'' as lack of consensus: a CQ was flagged as ambiguous if at least one annotator disagreed on suitability or explicitly commented on unclear meaning.

\noindent \paragraph{Relevance:} measures the degree to which a CQ aligns with the requirements expressed in the user story.
This assessment was performed by an LLM (\texttt{Gemini 2.5 Pro}), which was instructed to rate the relevance of each CQ on a 4-point Likert scale.
The scale was defined to capture different aspects of relevance from an ontology engineering perspective, where higher scores indicate closer alignment with explicitly stated or functionally necessary requirements.
Specifically, the scale points represented the following:
\textit{The CQ addresses a requirement explicitly stated in the user story - 4};
\textit{The CQ pertains to a requirement that, while not explicit, is inferable from the user story using domain knowledge and is functionally necessary for fulfilling the story's goals - 3};
\textit{The CQ is not directly inferable from the user story but holds some contextual relevance to the persona or their objectives within the story - 2};
\textit{The CQ introduces an additional requirement that is neither expressed in, nor inferable from, the user story and is not considered necessary - 1}.
A subset of these LLM judgments (12 CQs, with their expected ratings) were manually validated for prompt engineering.
We consider this relevance measure serves as a consistent proxy for expert judgment.

\begin{table}[t!]
\caption{Summary of readability measures used in this study. $|S|$, $|W|$, and $|Syl|$ denote the number of sentences, words, and syllables in the text, respectively. $|DC_{DW}|$ is the number of difficult words after excluding Dale-Chall's list of 3k common words.}
\label{tab:readabilityMeasures}
\begin{tabularx}{\textwidth}{|p{2.1cm}|p{2.9cm}|X|}
\hline
\textbf{Readability Measure}  & \textbf{Formula} & \textbf{Description} \\
\hline
Flesch-Kincaid Grade Level ({\bf FKGL}) \cite{kincaid1975} & $11.8 \times (\frac{|Syl|}{|W|}) + 0.39 \times (\frac{|W|}{|S|}) - 15.59$ & Estimates the years of formal education (U.S. grade, from kindergarten/nursery to college) a person needs to understand a piece of writing on the first reading. \\
\hline
Dale-Chall Readability Index ({\bf DCR}) \cite{dale1948} & $0.1579 \times (\frac{|DC_{DW}|}{|W|} \times 100) + 0.0496 \times (\frac{|W|}{|S|})$ & 
Estimates the comprehension difficulty of text based on word familiarity and sentence structure (using a specific list of 3,000 familiar words that are easily understood by 4th-grade students).
\\
\hline
\end{tabularx}
\end{table}

\noindent \paragraph{Readability:}
Each CQ was assessed to gauge its ease of understanding.
As done in~\cite{RevOnt_2024}, a suite of established readability indices designed to capture different aspects of textual difficulty were initially computed for each CQ using the \texttt{textstat} Python library.
For the subsequent analysis and reporting presented in this paper, we selected the Flesch-Kincaid Grade Level (FKGL) and the Dale-Chall Readability Score (DCR) as representative readability features that estimate the reader education level required to understand the text.

This selection was informed by a correlation analysis on AskCQ, which indicated that these two indices largely account for the variance observed across the broader set of metrics initially considered.\footnote{The correlation analysis is provided in the supplementary material.}
The focus of the selected metrics is the estimated US grade level required for comprehension (FKGL) and the reliance on common vocabulary (DCR), as explained in Table~\ref{tab:readabilityMeasures}.
Generally, lower scores on such indices indicate better readability, suggesting that less specialist knowledge or lower formal educational attainment is required for comprehension.

It is worth noting that most readability formulae, including FKGL and DCR, were primarily developed and validated for continuous prose rather than short, interrogative sentences like CQs.
Consequently, we interpret the resulting scores  as \textit{comparative indicators} of readability across the different CQ formulation approaches, rather than absolute measures of 
understandability for individual CQs.

\noindent \paragraph{Complexity:} of CQs can vary significantly, impacting modelling effort, potential ambiguity, and the intricacy of subsequent data querying.
To gain a quantitative understanding of these variations, we employ a multi-faceted approach that characterises each CQ along three distinct dimensions: semantic requirement, surface linguistic, and syntactic structural complexity.

\begin{itemize}
    \item \textbf{Requirement complexity.} 
    This dimension quantifies the complexity inherent in the emergence of \textit{ontological primitives}, potentially required to represent and retrieve the necessary information. The analysis focuses on the ``semantic payload'' of the CQ by identifying the fundamental ontological primitives implicitly or explicitly demanded.
    This is done by identifying the distinct \texttt{Concepts} (classes/entity types, e.g., \texttt{Item}, \texttt{Artist}), \texttt{Properties} (data attributes, e.g., \texttt{name}, \texttt{duration}), \texttt{Relationships} (links between concepts, e.g., \texttt{isPartOf}, \texttt{usedBy}), and specific \texttt{Filters} (constraints on properties or class memberships, e.g., ``main description''). Additionally, the expected \texttt{Cardinality} (implying retrieval of a single instance, multiple instances, or an existence check) and any \texttt{Aggregation} requirements (like counting or averaging) are identified. These primitives are expected to reflect the anticipated richness and interconnectedness of the induced ontological model and the likely complexity of corresponding queries. The primitives are extracted by an LLM (\texttt{Gemini 2.5 Pro}) and aggregated through a scoring function that sums the number of occurrences for each metric ($c_1$ complexity).

    \item \textbf{Linguistic complexity.} 
    This dimension assesses complexity based on the surface features of the CQ's \textit{natural language expression}. It quantifies aspects of the question's wording, length, and basic grammatical composition using NLP methods for part-of-speech tagging and noun chunking.
    Metrics are extracted automatically via \texttt{spaCy}, and include counts of \texttt{Noun Phrases} (proxy for entity/argument density), \texttt{Verbs} (actions/states), \texttt{Prepositions} (often indicating relational phrases), \texttt{Coordinating Conjunctions} (suggesting logical combinations), and \texttt{Modifiers} (adjectives/adverbs, reflecting descriptive detail or potential filters). The primary \texttt{Interrogative Structure} (e.g., WH-retrieval, Boolean check, aggregation via ``How many'') is also identified. This analysis provides a measure of the requirement statement's textual elaboration.
    Similarly, a scoring function summing over the metrics count with equal weighting is used to obtain a scalar ($c_2$ complexity).

    \item \textbf{Syntactic complexity:} 
    This dimension delves into the \textit{grammatical structure and relational syntax} of the CQ via dependency parsing. The analysis quantifies structural complexity through metrics calculated by traversing the generated dependency tree for each CQ.
    These include the total \texttt{Node Count} (token length), the maximum dependency path length or \texttt{Tree Depth} (indicating syntactic nesting and long-distance dependencies), and the frequency of specific \texttt{Dependency Relation} types indicative of complex grammatical constructions i.e., \texttt{nsubj}, \texttt{dobj}, \texttt{prep}, \texttt{acl}, \texttt{relcl}, \texttt{conj}, \texttt{agent}, selected based on linguistic complexity heuristics from the Universal Dependency set \cite{de-marneffe-etal-2014-universal}. 
    All counts and depths are aggregated by sum ($c_3$ complexity).
\end{itemize}
In addition, we also compute the \textit{length} of each CQ in terms of the number of characters ($c_0$ complexity) as a coarse-grained indicator of verbosity.
Overall, we expect these four dimensions -- length, requirement, linguistic, and syntactic structure –- to provide complementary perspectives on CQ complexity.
A CQ might be semantically complex (e.g., requiring navigation of intricate partonomy or causality relations) yet linguistically simple (e.g., ``What caused this event?''), scoring high on requirement metrics but low on linguistic/syntactic ones.
Conversely, a CQ might be ontologically straightforward but phrased using complex sentence structures, scoring high on syntactic metrics but low on semantic ones.
%
Despite lacking a formal definition of CQ complexity and having some methodological limitations (e.g., reliance on scoring mechanisms and basic induction), our method aims to (i) capture aspects that are likely to correlate with CQ complexity and (ii) analyze how these metrics differ across the AskCQ sets.

\subsection{Semantic analysis of CQ sets via sentence embeddings}\label{ssec:embeddings-methodology}



We analyse the semantic properties of the CQ sets (RQ3) by computing we compute Sentence-BERT embeddings using the *all-MiniLM-L6-v2* model \cite{reimers-2019-sentence-bert}, which maps each CQ to a 384-dimensional semantic vector, in line with previous approaches \cite{alharbi-et-al2024:ekaw,RevOnt_2024}. We then quantify the semantic overlap between each pair of CQ sets (e.g., \csets{Set A}{Set B}). For each pair, (N\_A = |A|) and (N\_B = |B|) denote the number of CQs in each set; we measure:

\begin{itemize}
    \item \textbf{Centroid cosine similarity.} The cosine similarity $\cos(\mathbf{\bar{e}}_A, \mathbf{\bar{e}}_B)$ between the centroids $\mathbf{\bar{e}}_A$ and $\mathbf{\bar{e}}_B$ of Set A and Set B provides a measure of the overall alignment of their central semantic representation. A score closer to 1 indicates that the two sets are, on average, focused on similar concepts.

    \item \textbf{Coverage analysis.} We measure how well one set covers the semantic content of another. This is performed in both directions, i.e. for the coverage of Set A by Set B (Set A $\leftarrow$ Set B) we compute:
    \begin{itemize}
        \item \textbf{Mean Maximum Similarity (MMS).} For each CQ embedding $\mathbf{e}_{A,i}$ in Set A, its maximum cosine similarity to any CQ embedding in Set B, $s_{A_i \rightarrow B} = \max_{j} \cos(\mathbf{e}_{A,i}, \mathbf{e}_{B,j})$, is identified. The mean of these $s_{A_i \rightarrow B}$ scores (and the standard deviation) indicate, on average, how well each CQ in Set A is semantically represented by its closest counterpart in Set B. A higher mean suggests stronger semantic parallels offered by Set B.
        \item \textbf{Set coverage and novelty.} The percentage of CQs in Set A for which $s_{A_i \rightarrow B} \geq \tau$ (where $\tau$ is a pre-defined similarity threshold) was calculated. This quantifies the proportion of Set A's semantic content considered adequately ``explained'' or represented by Set B. 
        Consequently, the percentage novelty represents the proportion of Set A that introduces semantic content not found (or not closely matched via $\tau$) in Set B.
    \end{itemize}
    The same metrics were computed for the coverage of Set B by Set A.

    \item \textbf{Bidirectional coverage:} This symmetric metric quantifies the overall mutual semantic overlap. It is calculated as $\frac{N_{A \rightarrow B}^{\textnormal{cov}} + N_{B \rightarrow A}^{\textnormal{cov}}}{N_A + N_B}$, where $N_{A \rightarrow B}^{\textnormal{cov}}$ is the number of CQs in Set A covered by Set B (i.e., $s_{A_i \rightarrow B} \geq \tau$), and $N_{B \rightarrow A}^{\textnormal{cov}}$ is the number of CQs in Set B covered by Set A. Hence, a higher percentage indicates greater shared conceptual space between the two sets.
\end{itemize}

Overall, these measures facilitate a multi-faceted comparative analysis of CQs generated from a controlled source.
Although relevance metrics and general-purpose embeddings rely on parameter choices and have inherent limitations, these factors affect the results comparably across all CQ sets.
Hence, we frame these computational metrics as comparative indicators of the trends characterising each generation approach rather than as absolute values.
\section{Results}\label{sec:results}

\begin{table}[t]
\caption{Expert evaluation results per CQ set: number of CQs (\#), proportion of commented CQs, suitability score, and proportion deemed suitable by majority vote.}
\label{tab:expert_evaluation}
\setlength{\tabcolsep}{8pt}
\centering
\begin{tabular}{@{}l c c c c@{}}
\toprule
\textbf{CQ Set} 
& \textbf{\#} 
& \textbf{Commented} (\%) & \textbf{Score} ($\mu \pm \sigma$) & \textbf{Score $>0$ CQs} (\%) \\
\midrule
HA-1           & 44 & $0.27$ & $2.39 \pm 1.26$ & $0.91$ \\
HA-2           & 54 & $0.19$ & $2.87 \pm 0.62$ & $0.98$ \\
Pattern        & 38 & $0.37$ & $0.11 \pm 2.12$ & $0.50$ \\
GPT            & 26 & $0.35$ & $1.85 \pm 1.52$ & $0.85$ \\
Gemini         & 42 & $0.31$ & $1.52 \pm 1.88$ & $0.67$ \\
\bottomrule
\end{tabular}
\end{table}

This section presents the results of our comparative analysis of CQ elicitation approaches.
The findings are structured according to our research questions (c.f. Section~\ref{sec:methodology}).
We first report the outcomes of the expert evaluation regarding CQ suitability (RQ1), followed by the comparative analysis of CQ features (RQ2), and conclude with the results of the semantic analysis exploring diversity and overlap between the generated CQ sets (RQ3).

In all experiments, LLMs from OpenAI and Google Cloud were prompted via their respective APIs using a configuration designed to maximise reproducibility.
The parameters were fixed as follows: $\text{temperature}=0$, $\text{frequency penalty}=0$, $\text{presence penalty}=0$, and $\text{seed}=46$.
This same configuration was used to generate the LLM-based CQ sets included in AskCQ (\texttt{GPT4.1} and \texttt{Gemini 2.5 Pro}; see Section~\ref{sec:methodology}).
All code and materials needed to reproduce our experiments are publicly available on GitHub under an MIT license.

\subsection{Expert evaluation results (RQ1)}\label{ssec:manual_results}

We report the CQ suitability score, computed by summing the independent ratings from the three expert annotators for each CQ.
This score ranges from -3 (unanimous rejection) to +3 (unanimous acceptance), where a score greater than 0 signifies acceptance by the majority of annotators.
To assess the reliability of this evaluation, we calculated the inter-annotator agreement using Fleiss' Kappa, yielding a coefficient of $\kappa = 0.35$, which indicates a fair agreement among the evaluators~\cite{landis1977measurement}.
The aggregated results are reported in Table~\ref{tab:expert_evaluation}.

Overall, the human-annotated sets received the highest suitability ratings from the expert evaluators.
\texttt{HA-2} achieved the highest mean score ($2.87 \pm 0.62$) with an acceptance rate of 98\%.
\texttt{HA-1} also performed strongly, with a mean score of $2.39 \pm 1.26$ and a 91\% acceptance rate.
These manually generated sets also received the lowest proportion of comments from evaluators (19\% for \texttt{HA-2}, 27\% for \texttt{HA-1}), suggesting fewer perceived issues compared to other methods.

LLM-generated CQs showed moderate suitability.
\texttt{GPT} obtained a mean score of $1.85 \pm 1.52$ with 85\% acceptance, while \texttt{Gemini} scored lower ($1.52 \pm 1.88$) with a 67\% acceptance rate.
The proportion of commented CQs for the LLM sets (35\% for \texttt{GPT}, 31\% for \texttt{Gemini}) was higher than for the human sets.

Notably, pattern-based CQs recorded the highest proportion of comments (37\%) and the lowest mean suitability score ($0.11$) coupled with the highest variability ($\sigma = 2.12$).
Hence, these CQs had the lowest acceptance rate (50\%).


\begin{table}[t]
\centering
\caption{Overview of CQ set features. Ambiguity is the percentage of CQs identified as ambiguous. Relevance reports the proportion of CQs rated as domain-relevant (score 3), as the remaining CQs received the highest relevance score (4). FKGL and DCR denote the Flesch-Kincaid Grade Level and Dale-Chall Readability Index, respectively.}
\label{tab:combined_summary}
\begin{tabular}{@{}l c c c c@{}}
\toprule
\textbf{CQ Set} 
& \textbf{Ambiguity} (\%) & \textbf{Relevance} (\% Score 3)
& \textbf{FKGL} ($\mu \pm \sigma$) & \textbf{DCR} ($\mu \pm \sigma$)\\
\midrule
HA-1           & 20.5 & 18.2 & $5.63 \pm 2.80$ & $9.59 \pm 1.62$ \\
HA-2           &  3.7 & 27.8 & $6.88 \pm 3.42$ & $8.76 \pm 2.00$ \\
Pattern        & 13.2 & 13.5 & $7.66 \pm 2.81$ & $10.94 \pm 2.63$ \\
GPT            &  3.8 & 12.0 & $11.64 \pm 2.69$ & $12.67 \pm 1.89$ \\
Gemini         & 14.3 &  4.8 & $9.72 \pm 2.67$ & $12.90 \pm 2.64$ \\
\bottomrule
\end{tabular}
\end{table}

\begin{table}[t]
\centering
\small
\caption{Mean and standard deviation of complexity measures per CQ set. $c_0$ denotes the number of characters in a CQ; $c_1$ quantifies the richness of ontological primitives; $c_2$ and $c_3$ measure surface-level and structural complexity, respectively.}
\label{tab:complexity_measures}
\begin{tabular}{l c c c c}
\toprule
& \multicolumn{4}{c}{\textbf{Complexity Facets}} \\
\cmidrule(lr){2-5}
\textbf{CQ Set} & \textbf{Length ($c_0$)} & \textbf{Requirement ($c_1$)} & \textbf{Lexical ($c_2$)} & \textbf{Syntactic ($c_3$)} \\
\midrule
HA-1    & $42.57 \pm 12.94$ & $4.52 \pm 1.50$ & $6.96 \pm 1.89$ & $16.76 \pm 5.27$ \\
HA-2    & $46.93 \pm 12.01$ & $4.17 \pm 1.09$ & $7.10 \pm 1.43$ & $18.04 \pm 3.87$ \\
Pattern & $51.69 \pm 15.53$ & $4.94 \pm 1.34$ & $6.76 \pm 1.47$ & $16.79 \pm 4.03$ \\
GPT     & $111.15 \pm 17.18$ & $8.12 \pm 2.14$ & $14.25 \pm 2.52$ & $37.91 \pm 7.20$ \\
Gemini  & $93.50 \pm 27.52$ & $5.60 \pm 2.40$ & $11.36 \pm 2.87$ & $31.96 \pm 9.06$ \\
\bottomrule
\end{tabular}
\end{table}

\subsection{Computational analysis of CQ features (RQ2)}\label{ssec:auto_results}

The proportion of ambiguous CQs and their relevance scores are outlined in Table~\ref{tab:combined_summary}.
Regarding \textbf{ambiguity},
\texttt{HA-2} set exhibited the lowest proportion (3.7\%), closely followed by \texttt{GPT} (3.8\%).
In contrast, \texttt{HA-1} had the highest proportion of CQs considered ambiguous (20.5\%), whilst the \texttt{Pattern} (13.2\%) and \texttt{Gemini} (14.3\%) sets showed moderate levels of ambiguity.

For \textbf{relevance}, Table~\ref{tab:combined_summary} also reports the proportion of CQs assigned a score of 3 on a 4-point Likert scale, which signifies CQs pertaining to requirements that are \textit{inferable} from the user story using domain knowledge, and are functionally necessary for fulfilling the story's goals; whilst all other CQs in this analysis achieved the maximum score of 4 (addressing explicitly stated requirements).
Notably, human-annotated sets, particularly \texttt{HA-2} (27.8\% score 3) and \texttt{HA-1} (18.2\% score 3), had the highest proportions of these \textit{inferential CQs}.
This suggests that the ontology engineers utilised their domain expertise to elicit questions that, whilst not directly stated, capture key functional needs, such as ``\textit{What is the family of an instrument?}'' or ``\textit{What is the format of each multimedia file?}'' (both of which received a relevance score of 3).
Conversely, \texttt{Gemini} had the lowest proportion of inferential (score 3) CQs (4.8\%), indicating that its generated questions adhered most closely to the requirements that are explicitly stated in the user story.
The \texttt{Pattern} (13.5\%) and \texttt{GPT} (12.0\%) sets presented an intermediate proportion of such inferential CQs.


\begin{figure}[t]
    \centering
    \includegraphics[width=0.70\linewidth]{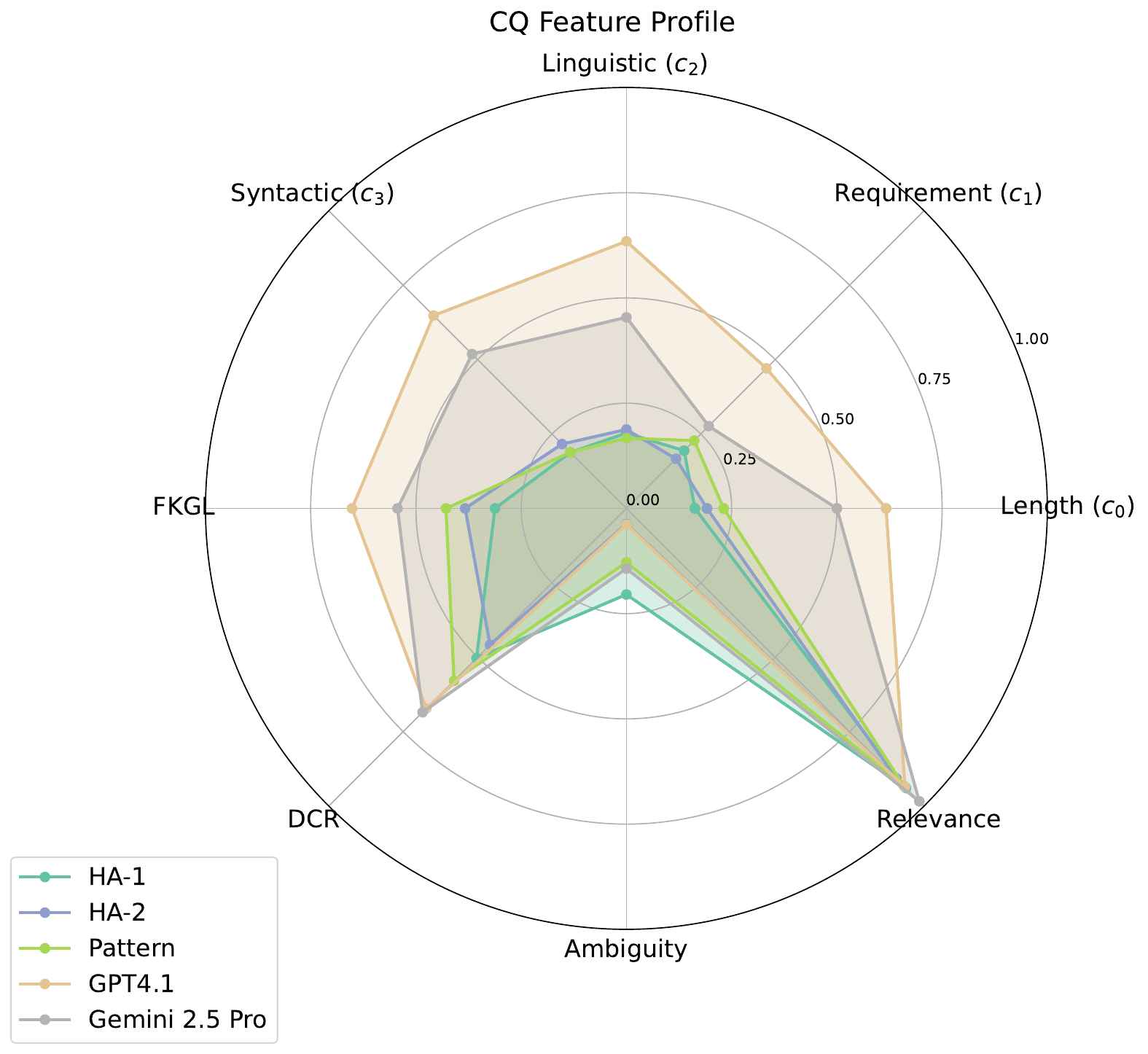}
    \caption{Min-max normalised CQ feature profile per elicitation approach/set in AskCQ.}
    \label{fig:feature-profile}
\end{figure}

The computational \textbf{readability} analysis, summarised in Table~\ref{tab:combined_summary}, reveals distinct patterns across the different CQ formulation approaches.
Manual CQs exhibited the highest readability; \texttt{HA-1} consistently achieved the lowest scores across most indices (e.g., Flesch-Kincaid Grade Level, FKGL, $5.63 \pm 2.80$), suggesting a reading level appropriate for early secondary school years.
\texttt{HA-2} scored slightly higher but remained highly readable (FKGL $6.88 \pm 3.42$), with the lowest Dale-Chall Readability score (DCR $8.76 \pm 2.00)$.
Pattern-based CQs presented slightly reduced readability compared to the manual sets (e.g., FKGL $7.66 \pm 2.81$).
In contrast, LLM-generated CQ were markedly less readable, requiring comprehension levels estimated at late secondary school or early university stages.
\texttt{GPT} yielded the lowest readability at the syllable level (FKGL $11.64 \pm 2.69$); and Gemini produced slightly more readable CQs than \texttt{GPT} (FKGL $9.72 \pm 2.67$), although the higher Dale-Chall Readability score (DCR $12.90 \pm 2.64$) potentially indicates more complex or less common vocabulary usage.

Finally, the analysis of CQ \textbf{complexity} across four dimensions -- character length ($c_0$), requirement ($c_1$), linguistic ($c_2$), and syntactic ($c_3$) --revealed significant variations based on elicitation approach (Table~\ref{tab:complexity_measures}).
Overall, LLM-generated CQs were consistently and markedly more complex: they were substantially longer, averaging over 90 characters and roughly double the length of CQs from other methods (which averaged 42--52 characters); \texttt{GPT} produced the longest CQs on average ($111.15 \pm 17.18$).
Similarly, requirement complexity ($c_1$), reflecting the richness of emerging ontological primitives, was significantly higher for LLMs.
\texttt{GPT} CQs exhibited the highest requirement complexity ($8.12 \pm 2.14$), suggesting they demand a more extensive set of concepts, properties, and relations compared to the manual and \texttt{Pattern} sets, whose scores clustered in the 4--5 range.
Within the latter group, \texttt{HA-2} CQs showed the lowest requirement complexity ($4.17 \pm 1.09$).
This trend of increased complexity for LLM-generated CQs extended to linguistic ($c_2$) and syntactic ($c_3$) measures, with both \texttt{GPT} and \texttt{Gemini} demonstrated considerably more elaborate phrasing and complex grammatical structures compared to the manual sets.
To summarise, these results indicate that the chosen elicitation approach fundamentally shapes multiple facets of the resulting complexity profile, with generative models yielding notably more complex CQs in this study.
Figure~\ref{fig:feature-profile} consolidates our findings, showing distinct feature profiles from the CQs generated by each elicitation approach.

\subsection{Semantic analysis of CQ sets (RQ3)}
\label{subsec:semantic_results}
We report the results of the semantic analysis conducted on the CQ sets using sentence-level embeddings (Section~\ref{ssec:embeddings-methodology}) with a similarity threshold of $\tau=0.75$.

The pairwise comparison results, detailed in Table~\ref{tab:pairwise_comparison}, reveal insights into the semantic overlap between CQ sets generated by different approaches.
Overall, the cosine similarities between the centroids of most pairs are relatively high (typically ranging from $0.61$ to $0.85$).
For instance, \csets{HA-1}{GPT} showed the highest centroid similarity ($0.85$), followed by \csets{HA-2}{Pattern} ($0.84$) and \csets{HA-1}{Pattern} ($0.83$). 
This suggests that, at a high level, all sets tend to address the same core thematic area defined by the user story.
The lowest centroid similarities were observed in comparisons involving \texttt{Gemini} (e.g., $0.61$ with \texttt{HA-2}), indicating its central theme might be slightly more distinct than the other sets.

Despite these relatively high centroid similarities, the specific semantic coverage between sets is low, denoting high degrees of novelty, i.e. a high number of CQs not previously generated.
The percentage of CQs in one set covered by another (i.e. having a CQ in the other set with similarity $\geq 0.75$) is consistently below $21\%$, and often below $10\%$.
Between the two human annotators (\csets{HA-1}{HA-2}), who shared a high centroid similarity ($0.82$), \texttt{HA-2} covered $20.5\%$ of \texttt{HA-1}'s CQs, and HA-1 covered $11.1\%$ of \texttt{HA-2}'s CQs (\texttt{HA-2} has 10 more CQs than \texttt{HA-1}), yielding a bidirectional coverage proportion of $15.3\%$.
These were the highest levels of directional and bidirectional coverage recorded across the sets.
In fact, the bidirectional coverage proportion, reflecting mutual overlap, remained low for all other pairs, frequently falling below $5\%$ and reaching $0\%$ for comparisons involving \texttt{Gemini} with human annotators.

This pattern of low coverage and high novelty extends across other comparisons.
For instance, \texttt{HA-1} showed only $9.1\%$ coverage by GPT and $0.0\%$ coverage by \texttt{Gemini}.
Similarly, \texttt{Gemini} exhibited $0\%$ coverage by both \texttt{HA-1} and \texttt{HA-2}, and very low coverage by Pattern ($2.4\%$) and GPT ($2.4\%$).
This highlights Gemini as producing a particularly distinct set of CQs at the chosen similarity threshold, despite its centroid similarities.
The Mean Maximum Similarity scores, even when coverage is zero (e.g., \csets{HA-1}{Gemini} MMS being $0.53 \pm 0.11$), indicate that while the closest CQs do not meet the similarity threshold ($\tau=0.75$), some underlying semantic relation (albeit weaker) exists.

In conclusion, the semantic analysis revealed several key findings.
Firstly, while thematically consistent, each elicitation approach tends to produce a set of CQs that is largely unique in its specific semantic expression.
Notably, \texttt{Gemini} CQs (18 more than GPT) recorded the lowest overlap with those produced by human annotators and the pattern-based method.
Notably, the human-annotated CQ sets show the highest directional (20.5\%) and bidirectional coverage (15.3\%) and MMS, demonstrating higher alignment in the identification of requirements.
The same behaviour is not observed among LLMs as CQ elicitators.

\begin{table}[t]
\centering
\small 
\setlength{\tabcolsep}{1pt} 
\caption{Pairwise semantic comparison of CQ sets over embeddings (similarity threshold $\tau=0.75$). Each row compares two sets (\csets{Set1}{Set2}) based on: centroid cosine similarity (overall thematic alignment), directional coverage (\% of Set1's CQs represented by Set2 and vice versa), and MMS (mean of maximum similarities per CQ). Bidirectional coverage measures the overall shared semantic space among sets (c.f. Section\ref{ssec:embeddings-methodology}).}
\label{tab:pairwise_comparison}
\begin{tabular}{@{}l c ccc ccc c@{}}
\toprule
\textbf{Comparison} & \textbf{Centroid} & \multicolumn{2}{c}{\textbf{Set1 $\leftarrow$ Set2}} & & \multicolumn{2}{c}{\textbf{Set1 $\rightarrow$ Set2}} & & \textbf{BiDirect.} \\
\cmidrule(lr){3-4} \cmidrule(lr){6-7}
(Set1 $\leftrightarrow$ Set2) & \textbf{Sim.} & \textbf{Cov.(\%)} & \textbf{MMS} && \textbf{Cov.(\%)} & \textbf{MMS} && \textbf{Cov.(\%)} \\
\midrule
HA-1, HA-2                 & 0.82 & \textbf{20.5} & 0.62 $\pm$ 0.15 && 11.1 & 0.58 $\pm$ 0.15 && \textbf{15.3} \\
HA-1, Pattern              & 0.83 & 15.9 & 0.61 $\pm$ 0.15 && \textbf{13.2} & 0.57 $\pm$ 0.16 && 14.6 \\
HA-1, GPT                  & 0.85 &  9.1 & 0.56 $\pm$ 0.13 && 11.5 & 0.60 $\pm$ 0.12 && 10.0 \\
HA-1, Gemini               & 0.73 &  \underline{0.0} & 0.53 $\pm$ 0.11 &&  \underline{0.0} & 0.53 $\pm$ 0.11 &&  \underline{0.0} \\
HA-2, Pattern              & 0.84 &  9.3 & 0.57 $\pm$ 0.15 && \textbf{13.2} & 0.59 $\pm$ 0.15 && 10.9 \\
HA-2, GPT                  & 0.74 &  1.9 & 0.48 $\pm$ 0.11 &&  3.8 & 0.55 $\pm$ 0.10 &&  2.5 \\
HA-2, Gemini               & 0.61 &  \underline{0.0} & 0.46 $\pm$ 0.13 &&  \underline{0.0} & 0.51 $\pm$ 0.12 &&  \underline{0.0} \\
Pattern, GPT               & 0.77 &  5.3 & 0.49 $\pm$ 0.16 &&  7.7 & 0.56 $\pm$ 0.14 &&  6.2 \\
Pattern, Gemini            & 0.68 &  2.6 & 0.55 $\pm$ 0.11 &&  2.4 & 0.53 $\pm$ 0.12 &&  2.5 \\
GPT, Gemini                & 0.80 &  3.8 & 0.61 $\pm$ 0.13 &&  2.4 & 0.57 $\pm$ 0.11 &&  2.9 \\
\bottomrule
\end{tabular}

\end{table}

\section{Discussion}\label{sec:discussion}

Our comparative analysis of CQ elicitation approaches --manual, pattern-based, and LLM-based -- on AskCQ produced significant findings.
The expert evaluation (\textbf{RQ1}) revealed a clear preference for CQs formulated by ontology engineers (\texttt{HA-1}, \texttt{HA-2}).
These two sets achieved the highest suitability scores and acceptance rates (over 90\%), and notably, received the least number of comments, suggesting that they had fewer perceived issues by the expert evaluators (Table~\ref{tab:expert_evaluation}).
This preference aligns with the findings from our analysis of CQ features (\textbf{RQ2}).
Human-generated CQs consistently demonstrated higher readability (lower Flesch-Kincaid Grade Level and Dale-Chall scores) and lower complexity across all four dimensions assessed: character length, requirement, linguistic, and syntactic complexity (Tables~\ref{tab:combined_summary} and \ref{tab:complexity_measures}).
In contrast, LLM-generated CQs (\texttt{GPT} and \texttt{Gemini}) were generally rated as less suitable, were significantly less readable, and exhibited substantially higher complexity.
Pattern-based CQs occupied a middle ground across the CQ features, but were found less suitable.
Ambiguity could not be ascribed to any specific method (Table~\ref{tab:combined_summary}), and indeed, all methods generated comments that necessitated discussion among the annotators.
Interestingly, the highest ambiguity was recorded for the human-annotated set \texttt{HA-1}, followed by \texttt{Gemini} and \texttt{Pattern}.
Nevertheless, 66\% of \texttt{HA-1}'s ambiguous CQs were resolved after a discussion phase with the annotators, as this set recorded a 91\% suitability score.
Instead, only 33\% of ambiguous CQs from \texttt{Gemini} were resolved as suitable CQs after discussion.
A crucial observation regarding relevance was that human CQs, particularly from \texttt{HA-2}, showed a higher proportion of CQs with \textit{inferential relevance} (Score 3) -- addressing requirements not explicitly stated in the user story but functionally necessary for its goals.
This points to the role of domain knowledge or ontology engineering (OE) experience in shaping CQs.

The semantic analysis of CQ sets (\textbf{RQ3}) provided further insights.
While all CQ sets demonstrated thematic alignment with the user story, as indicated by generally high centroid similarities (Table~\ref{tab:pairwise_comparison}), the semantic overlap at a granular CQ level was low.
The highest bidirectional coverage (15.3\%) was observed between the two human annotators (\texttt{HA-1} and \texttt{HA-2}), suggesting that while their formulation styles might differ, experienced engineers tend to converge on a core set of shared requirements.
Conversely, the CQs generated by the LLMs exhibited very little semantic overlap with each other (e.g., 2.9\% bidirectional coverage between \texttt{GPT} and \texttt{Gemini}) and often negligible coverage (even 0\%) when compared to human-generated CQs.


Overall, these findings corroborate the need of human expertise to eliciting high-quality requirements.
Relying solely on automated methods, especially in their current off-the-shelf form, might result in CQs that are difficult to understand, overly complex, or miss key implicit requirements.
The inferential relevance of CQs further highlights the value of expert knowledge and experience, which current automated methods still struggle to replicate.
The low semantic overlap among LLM-generated sets, and between LLMs and human experts, also appears concerning.
If different LLMs, or even the same LLM with different prompt, produce vastly different sets of CQs for the same input, this could introduce inconsistencies and gaps if used as the sole basis for OE.
This suggests that a hybrid approach may be most beneficial: LLMs could assist in initial brainstorming, but human refinement and augmentation are crucial for ensuring clarity, conciseness, relevance (especially inferential), and overall suitability.

While the use LLMs for CQ generation is a promising research direction and a growing trend~\cite{alharbi2024SAC,RevOnt_2024,rebboud2024_ESWC,Bohui2025}, our comparative results suggest that the task requires further investigation to ensure consistent output quality.
Specifically, the variability observed in our study underscores the necessity for clear instructions and carefully designed prompts regarding how to formulate CQs.
In this context the characteristics of highly-rated human CQs identified in this study
can serve as desiderata for improving automated CQ generation (e.g., via prompt engineering \cite{Bohui2025}).
For instance, instructing LLMs with examples of well-formed CQs, or explicitly prompting them to consider specific properties and a desired level of semantic overlap with a controlled set of requirements, might yield outputs that are more aligned with ontology engineers.
This also raises the question of what constitutes an ``optimal sets'' of CQs.
While high overlap might indicate consensus, a degree of diversity could ensure broader coverage.
Future work could explore ensemble methods for LLM-generated CQs or techniques to guide LLMs towards a more aligned, yet comprehensive, set of requirements.

Although the study’s results are significant, they are based on a single user story in the cultural heritage domain. Applying the approach to other domains or source materials may introduce confounding variables, making it difficult to attribute differences in CQs to the generation method rather than the source text. Therefore, its generalisability to other domains and requirement types requires further investigation.

The use of LLMs for assessing certain CQ features (relevance, components of complexity), though guided by careful prompt engineering and some manual validation, is subject to the inherent limitations of current LLM capabilities.
A similar consideration applies to the complexity metrics, which, while providing comparative insights, are proxies and do not constitute a formal, exhaustive definition of CQ complexity.
Hence, the value of both the LLM-based feature assessments and our complexity proxies lies in their capacity to support comparative analysis across CQ sets, rather than in the absolute range of their scores.

\section{Conclusions}\label{sec:conclusions}

The effective formulation of Competency Questions (CQs) is central for ontology engineering (OE), yet systematic comparisons of elicitation methods are scarce.
This paper presented an empirical comparative evaluation of manual, pattern-based, and LLM-driven CQ generation, introducing AskCQ, a novel multi-annotator dataset, and a multi-faceted evaluation framework.
Our findings highlight that CQs manually crafted by ontology engineers demonstrated the highest suitability for OE, due to achieving better readability, lower complexity, and uniquely capturing inferential requirements (implicit functional requirements) essential for robust ontology design.
While LLMs produce relevant and thematically coherent outputs, the resulting CQs exhibited higher complexity, lower readability, and their semantic coverage, though broad, showed limited overlap with human-generated CQs and amongst each other.

These results underscore that human expertise still remains central for producing effective CQs for OE.
Future work will extend this study to other user stories and domains to assess the generalisability of our findings.
Crucially, our insights on CQ characteristics and limitations of current automated approaches can be leveraged to directly inform and improve their elicitation methods, aiming to better align their outputs with the desiderata of ontology engineers.


\subsubsection*{Supplemental Material Statement} All data and code to reproduce our experiments are available at 
(\url{https://github.com/KE-UniLiv/askcq}).
%
%
%
\bibliographystyle{splncs04}
\bibliography{references.bib}

\end{document}